\documentclass[sigconf]{acmart}

\usepackage{times}
\usepackage{latexsym}
\usepackage{array}
\usepackage{arydshln}
\usepackage{booktabs}
\usepackage{graphicx}
\usepackage{amsmath}
\usepackage{algorithm}
\usepackage{algpseudocode}
\usepackage{multirow}
\usepackage{multicol}
\usepackage{longtable}

\usepackage[T1]{fontenc}
\usepackage[utf8]{inputenc}
\usepackage{microtype}
\usepackage{graphicx}

\copyrightyear{2025} 
\acmYear{2025} 
\setcopyright{cc}
\setcctype{by}
\acmConference[CIKM '25]{Proceedings of the 34th ACM International Conference on Information and Knowledge Management}{November 10--14, 2025}{Seoul, Republic of Korea}
\acmBooktitle{Proceedings of the 34th ACM International Conference on Information and Knowledge Management (CIKM '25), November 10--14, 2025, Seoul, Republic of Korea}\acmDOI{10.1145/3746252.3760855}
\acmISBN{979-8-4007-2040-6/2025/11}

\acmSubmissionID{csp3860}

\begin{document}

\title[Exploring Reasoning-Infused Text Embedding with LLMs for Zero-Shot Dense Retrieval]{Exploring Reasoning-Infused Text Embedding with Large Language Models for Zero-Shot Dense Retrieval}

% \author{Yuxiang Liu}
% \authornote{Correspondence author. Work done when doing an internship in Amazon.}
% \email{yuxiang@illinois.edu}
% \orcid{0009-0004-2015-4848}
% \affiliation{%

\author{Yuxiang Liu}
\authornote{Corresponding author. Work was done while Yuxiang was interning at Amazon.}
\affiliation{%
  \institution{University of Illinois at Urbana-Champaign}
  \city{Urbana}
  \country{USA}
  }
\email{yuxiang@illinois.edu}

\author{Tian Wang}
\affiliation{%
  \institution{Amazon}
  \city{Palo Alto}
  \country{USA}
}
\email{wangtan@amazon.com}

\author{Gourab Kundu}
\affiliation{%
 \institution{Amazon}
 \city{New York}
 \country{USA}
 }
\email{gkundu@amazon.com}

\author{Tianyu Cao}
\affiliation{%
  \institution{Amazon}
  \city{Palo Alto}
  \country{USA}
  }
\email{caoty@amazon.com}

\author{Guang Cheng}
\affiliation{%
  \institution{Amazon}
  \city{Los Angeles}
  \country{USA}
  }
\email{amazoncg@amazon.com}

\author{Zhen Ge}
\affiliation{%
  \institution{Amazon}
  \city{New York}
  \country{USA}
  }
\email{zge@amazon.com}

\author{Jianshu Chen}
\affiliation{%
  \institution{Amazon}
  \city{Palo Alto}
  \country{USA}
  }
\email{jianshuc@amazon.com}

\author{Qingjun Cui}
\affiliation{%
  \institution{Amazon}
  \city{Palo Alto}
  \country{USA}
  }
\email{qingjunc@amazon.com}

\author{Trishul Chilimbi}
\affiliation{%
  \institution{Amazon}
  \city{Palo Alto}
  \country{USA}
  }
\email{trishulc@amazon.com}

\renewcommand{\shortauthors}{Liu, et al.}

\begin{abstract}
Transformer-based models such as BERT and E5 have significantly advanced text embedding by capturing rich contextual representations.
However, many complex real-world queries require sophisticated reasoning to retrieve relevant documents beyond surface-level lexical matching, where encoder-only retrievers often fall short. 
Decoder-only large language models (LLMs), known for their strong reasoning capabilities, offer a promising alternative. 
Despite this potential, existing LLM-based embedding methods primarily focus on contextual representation and do not fully exploit the reasoning strength of LLMs. 
To bridge this gap, we propose \textit{Reasoning-Infused Text Embedding} (\textbf{RITE}), a simple but effective approach that integrates logical reasoning into the text embedding process using generative LLMs. 
RITE builds upon existing language model embedding techniques by generating intermediate reasoning texts in the token space before computing embeddings, thereby enriching representations with inferential depth. 
Experimental results on \textit{BRIGHT}, a reasoning-intensive retrieval benchmark, demonstrate that RITE significantly enhances zero-shot retrieval performance across diverse domains, underscoring the effectiveness of incorporating reasoning into the embedding process.
\end{abstract}




\maketitle

\vspace{-0.1in}
\section{Introduction}\label{sec:intro}

Information retrieval is fundamental to numerous applications, including search~\cite{kobayashi2000information}, recommendation~\cite{isinkaye2015recommendation}, retrieval-augmented generation~\cite{DBLP:conf/acl/LiuC25, 10.1145/3627673.3679071}, and question answering~\cite{zhang-etal-2024-end}. Effective retrieval relies on accurately representing large-scale text corpora and efficiently mapping queries to relevant information, with text embeddings playing a crucial role in capturing semantic nuances~\cite{patil2023survey, 11010625}. 

Transformer-based encoders like BERT~\cite{devlin-etal-2019-bert} or E5~\cite{wang2022text} significantly advance embedding by leveraging contextual representations to improve semantic understanding. However, real-world queries often require sophisticated reasoning, inference, and an understanding of implicit relationships beyond surface-level token similarity. Encoder-only retrievers, despite their success, frequently struggle with complex reasoning tasks~\cite{xiao2024rar, DBLP:conf/emnlp/Liu0C23}, necessitating additional components such as pre-retrieval query augmentation~\cite{ma-etal-2023-query, wang2024blendfilter} or post-retrieval cross-document reasoning~\cite{yu2023chain, trivedi-etal-2023-interleaving} to bridge the gap. 

Meanwhile, Large language models (LLMs)~\cite{jiang2023mistral, achiam2023gpt, ji2025leveraging, luo2025cross, 11065413, yang2024hades, luo2025federated} exhibit strong reasoning capabilities that make them attractive for information retrieval~\cite{xu2024can}. Recent research has explored their use in text embeddings via direct extraction~\cite{zhuang2024promptreps}, distillation~\cite{lee2024gecko}, or fine-tuning~\cite{ma2023fine, wang-etal-2024-improving-text, behnamghader2024llm2vec, muennighoff2024generative}. However, these approaches primarily focus on encoding contextual representations while underutilizing LLMs' intrinsic reasoning abilities, which are crucial for interpreting complex queries~\cite{xu2024restful} and establishing deeper semantic relationships.

To address this gap, we propose \textit{Reasoning-Infused Text Embedding} (\textbf{RITE}), a novel approach that seamlessly integrates reasoning into the embedding process using a single generative model. Unlike traditional methods that decouple reasoning and embedding across multiple components, RITE employs an LLM to directly perform reasoning in the token space before embedding extraction. This unified approach enables richer semantic representations, particularly in cases where complex understanding is required.

Building upon prior language model embedding techniques such as Echo~\cite{springer2024repetition} and PromptReps~\cite{zhuang2024promptreps}, RITE demonstrates substantial improvements in zero-shot retrieval performance when evaluated on BRIGHT~\cite{su2024bright}, a reasoning-intensive retrieval benchmark. Our results highlight the advantages of reasoning-infused embeddings, suggesting that integrating inferential depth into language model embeddings can significantly enhance retrieval capabilities. Furthermore, this work underscores the potential for future advancements through fine-tuning and optimization strategies, paving the way for more robust and intelligent retrieval systems.

% Our exploration of RITE underscores the potential of reasoning-infused embeddings, suggesting that further research--particularly in fine-tuning scenarios--could yield even greater improvements. 
% \input{sections/sec2-related}
\vspace{-0.1in}
\section{Methodology}

We propose Reasoning-Infused Text Embedding (RITE) to enhance zero-shot dense retrieval by incorporating reasoning into embedding using LLMs. RITE extends the capabilities of existing zero-shot embedding methods, Echo~\cite{springer2024repetition} and PromptReps~\cite{zhuang2024promptreps}, by involving an explicit reasoning step before embedding extraction. 
% This additional step enriches query representations, ensuring a deeper semantic understanding for enhanced retrieval performance.

\subsection{Zero-Shot Language Model Embedding}\label{sec:LMEmbedding}

Zero-shot dense retrieval leverages LLM-generated embeddings to retrieve relevant documents without training. We build upon two existing zero-shot embedding methods: Echo~\cite{springer2024repetition} and PromptReps~\cite{zhuang2024promptreps}, both of which serve as fundamental components of RITE.

% \subsubsection{Echo}

\textbf{Echo}~\cite{springer2024repetition} addresses the inherent limitation of auto-regressive language models, which lack bidirectional contextual awareness in a single forward pass. Echo hypothesizes that repeating a text sequence enables the model to reinforce its understanding of the sentence structure, thereby improving representation quality. To operationalize this, Echo applies a self-repetition strategy and extracts the mean token representation of the second occurrence of an input text $x$. The embedding is generated using the prompt\footnote{Note that "query" in all prompts can be replaced with "passage" when generating document embeddings.}: \textit{Rewrite the query: $x$, rewritten query: $x$}.

% % \vspace{0.03in}
% \noindent\fbox{%
%     \centering
%     \parbox{0.98\linewidth}{%
%         \textit{Prompt}: Rewrite the query: $x$, rewritten query: $x$.
%     }%
% }
% % \vspace{0.01in}

% \subsubsection{PromptReps}

\textbf{PromptReps} (\textbf{PR})~\cite{zhuang2024promptreps} leverages the compressive power of LLMs by instructing them to summarize a query or document into a single-word representation. Inspired by prior work on constrained-text embeddings~\cite{jiang2023scaling, zhang2024simple}, PR generates an embedding by extracting the first output token from the prompt: \textit{Query: $x$. Use one most important word to represent the query in a retrieval task. Make sure your word is in lowercase. The word is: ``}.

% % \vspace{0.03in}
% \noindent\fbox{%
%     \centering
%     \parbox{0.98\linewidth}{%
%         \textit{Prompt}: Query: $x$. Use one most important word to represent the query in a retrieval task. Make sure your word is in lowercase. The word is: ``
%     }%
% }
% % \vspace{0.01in}

While Echo and PR generate high-quality embeddings, neither method explicitly incorporates reasoning, which is essential for complex queries requiring inference. To address this gap, RITE integrates an additional reasoning step, enriching query representations with deeper semantic and inferential understanding.

\subsection{Reasoning-Infused Text Embedding (RITE)}

\begin{figure}[tp!]
\vspace{-0.15in}
\caption{Model overview.}
% \vspace{-0.05in}
\centerline{\includegraphics[width=0.9\linewidth]{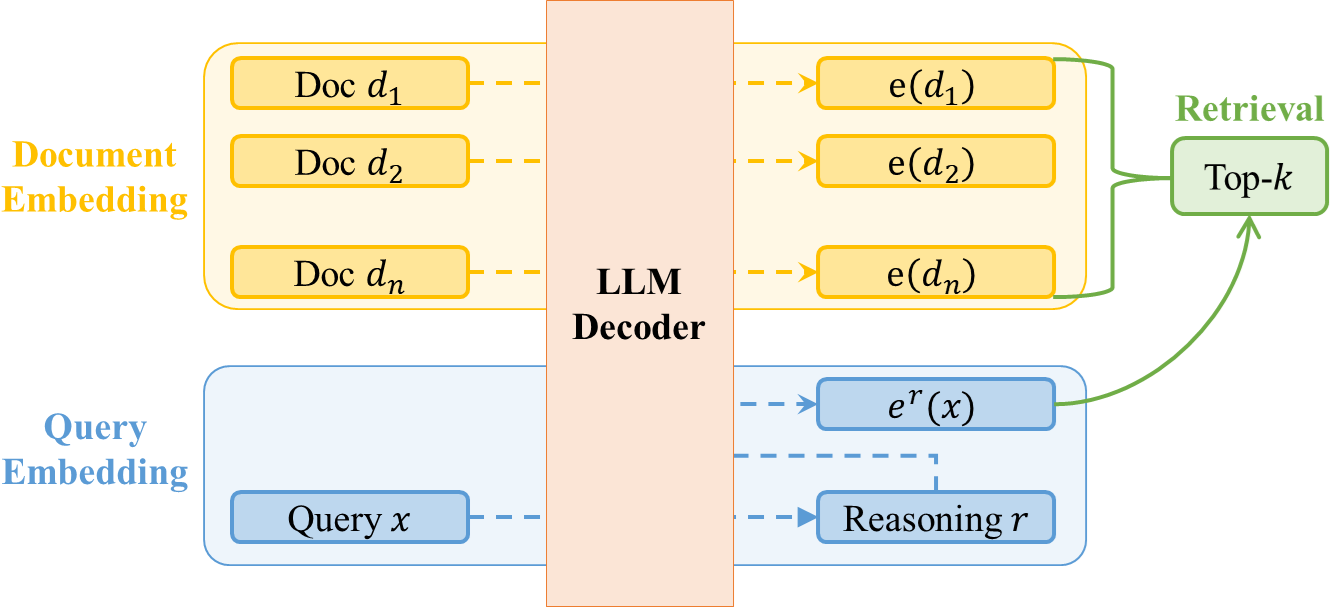}}
\vspace{-0.25in}
\label{fig:model}
\end{figure}

While Echo and PR effectively generate semantic-rich embeddings, they do not explicitly integrate logical reasoning, which is crucial for handling complex queries requiring inference. RITE (Figure~\ref{fig:model}) extends these methods by injecting a reasoning step before embedding extraction, enriching query representations with inferential depth. RITE improves query embeddings by introducing a reasoning stage before embedding extraction. It consists of three key steps: (1) \textbf{Reasoning Elicitation}: Generate an inferentially enriched reasoning text from the query. (2) \textbf{Embedding Extraction}: Apply Echo or PR on the reasoning-infused query to obtain a final embedding. (3) \textbf{Document Retrieval}: Compute similarity between the query and document embeddings to retrieve the most relevant documents.
% \begin{enumerate}
% \item Reasoning Elicitation: Generate an inferentially enriched reasoning text from the query.
% \item Embedding Extraction: Apply Echo or PromptReps on the reasoning-infused query to obtain a final embedding.
% \item Document Retrieval: Compute similarity between the query and document embeddings to retrieve the most relevant documents.
% \end{enumerate}
A high-level overview of the RITE framework is illustrated in Figure~\ref{fig:model}.

% To improve the retrieval of relevant documents for complex queries requiring deep reasoning, we introduce the Reasoning-Infused Text Embedding (RITE) method. Traditional encoder-based models, while effective for surface-level matching, often struggle with queries involving complex logical reasoning. In contrast, decoder-based large language models (LLMs), which excel in generating and processing contextual language, present a promising alternative for embedding generation by incorporating reasoning as an integral component of the retrieval process. RITE incorporates reasoning into the embedding generation process for zero-shot dense retrieval, where relevant documents are retrieved based on embeddings without prior task-specific training.

\subsubsection{Step 1: Eliciting Reasoning}

Given a query $x$, we first prompt the LLM to generate a reasoning text $r$ that provides additional contextual understanding, reformulation, or inferential expansion of the query. This is done using a reasoning-specific prompt $p^R$, resulting in the generated reasoning text:
% \vspace{-0.1in}
$$r=\text{LLM}(x, p^R)$$
% \vspace{-0.1in}

We experiment with the following three reasoning prompts designed to elicit deeper query understanding, which encourage LLM to reformulate the query in a more retrieval-effective manner.

% \vspace{0.1in}
\noindent\fbox{%
    \centering
    \parbox{0.98\linewidth}{%
        \textbf{Prompt 1:} Query: $x$. After thinking step by step, provide a better search query for search engine to answer the given question.
    }%
}
\noindent\fbox{%
    \centering
    \parbox{0.98\linewidth}{%
        \textbf{Prompt 2:} Query: $x$. Think step by step to reason about what is the essential problem of this question, and what should be included in the relevant documents. Make your response concise.
    }%
}
\noindent\fbox{%
    \centering
    \parbox{0.98\linewidth}{%
        \textbf{Prompt 3:} Query: $x$. After thinking step by step, provide a better search query for search engine to answer the given question, and identify what should be included in the relevant documents. Make your response concise.
    }%
}
% \vspace{0.1in}

\subsubsection{Step 2: Embedding Extraction}

Once reasoning $r$ is generated, it is prepended to the original query $x$ before embedding. RITE extends Echo and PR to integrate the reasoning, resulting in two variants: (1) RITE-Echo: Applies the Echo repetition strategy to the reasoning-enriched query. (2) RITE-PR: Extracts a compressed reasoning-infused representation using the PR method.
% \begin{itemize}
% \item RITE-Echo: Applies the Echo repetition strategy to the reasoning-enriched query.
% \item RITE-PromptReps (RITE-PR): Extracts a compressed reasoning-infused representation using the PromptReps method.
% \end{itemize}

These embedding strategies are formulated as follows:
(1) \textbf{RITE-Echo}: \textit{[Reasoning text] Rewrite the query: $x$, rewritten query: $x$}. (2) \textbf{RITE-PR}: \textit{Query: $x$. [Reasoning text] Use one most important word to represent the query in a retrieval task. Make sure your word is in lowercase. The word is: ``}.
% % \vspace{0.03in}
% \noindent\fbox{%
%     \centering
%     \parbox{0.98\linewidth}{%
%         \textbf{RITE-Echo:} [Reasoning text] Rewrite the query: $x$, rewritten query: $x$
%     }%
% }
% \noindent\fbox{%
%     \centering
%     \parbox{0.98\linewidth}{%
%         \textbf{RITE-PR:} Query: $x$. [Reasoning text] Use one most important word to represent the query in a retrieval task. Make sure your word is in lowercase. The word is: ``
%     }%
% }
% % \vspace{0.03in}
By incorporating explicit reasoning, RITE ensures that the embeddings capture not only semantic similarity but also the underlying logical structure of the query.

\subsubsection{Step 3: Document Retrieval}

For each document $d$ in the corpus $D$, an embedding is generated using either Echo or PR without reasoning augmentation. Retrieval is then performed by computing the cosine similarity between the reasoning-infused query embedding and the document embeddings:
% \vspace{-0.1in}
\begin{align*}
    \text{score}(x,d) = \text{cos}\left( \text{Embed}\left( \text{RITE}(x) \right), \text{Embed}(d) \right)
\end{align*}
% \vspace{-0.1in}

The top-$k$ documents with the highest similarity scores are retrieved as the most relevant results. By incorporating logical reasoning into the embedding process, RITE enables more robust retrieval of documents relevant to complex, inferential queries.
\vspace{-0.1in}
\section{Experimental Setup}

\begin{table*}[tb!]
\fontsize{7pt}{7pt}\selectfont
\centering
% \vspace{-0.05in}
  \caption{Retrieval performance (nDCG@10) with and without reasoning infusion.}
  \vspace{-0.15in}
\begin{tabular}{llrrrrrrr|rr|rrr|r}
    \toprule
    Backbones & Models & \textbf{Bio.} & \textbf{Earth.} & \textbf{Econ.} & \textbf{Psy.} & \textbf{Rob.} & \textbf{Stack.} & \textbf{Sus.} & \textbf{Leet.} & \textbf{Pony} & \textbf{AoPS} & \textbf{TheoQ.} & \textbf{TheoT.} & \textbf{Avg.} \\
    \midrule
    \multirow{4}*{Mistral 7B}
    & Echo & 10.8 & 7.5 & 7.4 & 8.9 & 0.1 & 3.3 & 7.6 & 13.4 & 15.1 & 2.7 & 17.4 & 1.8 & 8.0 \\
    & RITE-Echo & \underline{17.1} & \underline{11.5} & \underline{8.7} & \underline{10.9} & \underline{2.8} & \underline{4.6} & \underline{8.5} & \underline{15.3} & \underline{25.8} & \underline{3.1} & \underline{17.5} & \underline{2.3} & \underline{10.7} \\
    & PR & 4.5 & 6.1 & 3.0 & 5.3 & 1.1 & 2.2 & 2.7 & \underline{12.9} & 2.2 & 0.6 & \underline{16.9} & 2.4 & 5.0 \\
    & RITE-PR & \underline{15.1} & \underline{7.3} & \underline{8.2} & \underline{14.1} & \underline{2.3} & \underline{5.0} & \underline{5.9} & 12.1 & \underline{6.7} & \underline{1.6} & \underline{16.9} & \underline{7.5} & \underline{8.6} \\
    \midrule
    \multirow{4}*{LLaMA 3 8B}
    & Echo & 17.7 & 12.4 & 9.2 & 10.9 & 1.2 & 4.2 & 8.3 & 5.4 & 17.6 & 3.4 & \underline{17.7} & 3.3 & 9.3 \\
    & RITE-Echo & \underline{22.8} & \underline{16.9} & \underline{9.9} & \underline{12.7} & \underline{4.3} & \underline{5.3} & \underline{9.5} & \underline{16.8} & \underline{18.0} & \underline{4.5} & 16.7 & \underline{3.4} & \underline{11.7} \\
    & PR & 11.6 & 11.8 & 9.7 & 12.8 & 5.4 & 5.8 & \underline{11.2} & 16.4 & 2.9 & \underline{1.6} & 4.7 & 2.7 & 8.1 \\
    & RITE-PR & \underline{14.5} & \underline{13.7} & \underline{11.2} & \underline{17.8} & \underline{5.9} & \underline{6.6} & 10.5 & \underline{19.6} & \underline{7.5} & 1.5 & \underline{21.3} & \underline{11.5} & \underline{11.8} \\
    \bottomrule
  \end{tabular}
  \label{tab:results-main}
  \vspace{-0.05in}
\end{table*}

\begin{table*}[tb!]
\fontsize{7pt}{7pt}\selectfont
  \centering
  % \vspace{-0.05in}
  \caption{Echo-based retrieval performance -- model-generated (RITE-Echo) vs. oracle (GT-Echo) reasoning texts.}
  \vspace{-0.15in}
  \begin{tabular}{llrrrrrrr|rr|rrr}
  \toprule
    Backbones & Models & \textbf{Bio.} & \textbf{Earth.} & \textbf{Econ.} & \textbf{Psy.} & \textbf{Rob.} & \textbf{Stack.} & \textbf{Sus.} & \textbf{Leet.} & \textbf{Pony} & \textbf{AoPS} & \textbf{TheoQ.} & \textbf{TheoT.} \\
    \midrule
    \multirow{2}*{Mistral 7B}
    & RITE-Echo & 17.1 & 11.5 & 8.7 & 10.9 & \underline{2.8} & \underline{4.6} & 8.5 & \underline{15.3} & \underline{25.8} & 3.1 & 17.5 & 2.3 \\
    & GT-Echo & \underline{17.3} & \underline{11.9} & \underline{10.2} & \underline{11.4} & 2.7 & \underline{4.6} & \underline{9.4} & - & - & \underline{3.9} & \underline{18.4} & \underline{5.7} \\
    \midrule
    \multirow{2}*{LLaMA 3 8B}
    & RITE-Echo & 22.8 & 16.9 & 9.9 & 12.7 & 4.3 & 5.3 & 9.5 & \underline{16.8} & \underline{18.0} & 4.5 & 16.7 & 3.4 \\
    & GT-Echo & \underline{24.7} & \underline{19.1} & \underline{11.1} & \underline{16.6} & \underline{5.1} & \underline{5.4} & \underline{12.3} & - & - & \underline{5.4} & \underline{18.1} & \underline{8.4} \\
    \bottomrule
  \end{tabular}
  \label{tab:results-oracle-echo}
  \vspace{-0.05in}
\end{table*}

\begin{table*}[tb!]
\fontsize{7pt}{7pt}\selectfont
  \centering
  % \vspace{-0.05in}
  \caption{PR-based retrieval performance -- model-generated (RITE-Echo) vs. oracle (GT-Echo) reasoning texts.}
  \vspace{-0.15in}
  \begin{tabular}{llrrrrrrr|rr|rrr}
  \toprule
    Backbones & Models & \textbf{Bio.} & \textbf{Earth.} & \textbf{Econ.} & \textbf{Psy.} & \textbf{Rob.} & \textbf{Stack.} & \textbf{Sus.} & \textbf{Leet.} & \textbf{Pony} & \textbf{AoPS} & \textbf{TheoQ.} & \textbf{TheoT.} \\
    \midrule
    \multirow{2}*{Mistral 7B}
    & RITE-PR & \underline{15.1} & 7.3 & 8.2 & 14.1 & 2.3 & \underline{5.0} & 5.9 & \underline{12.1} & \underline{6.7} & 1.6 & 16.9 & 7.5 \\
    & GT-PR & 10.0 & \underline{9.9} & \underline{10.7} & \underline{15.1} & \underline{2.8} & 4.1 & \underline{9.5} & - & - & \underline{2.7} & \underline{18.6} & \underline{29.0} \\
    \midrule
    \multirow{2}*{LLaMA 3 8B}
    & RITE-PR & 14.5 & 13.7 & 11.2 & 17.8 & \underline{5.9} & \underline{6.6} & 10.5 & \underline{19.6} & \underline{7.5} & 1.5 & \underline{21.3} & 11.5 \\
    & GT-PR & \underline{16.2} & \underline{17.9} & \underline{17.8} & \underline{23.8} & 5.2 & 6.4 & \underline{14.3} & - & - & \underline{3.3} & 19.9 & \underline{33.5} \\
    \bottomrule
  \end{tabular}
  \label{tab:results-oracle-pr}
  \vspace{-0.05in}
\end{table*}

\subsection{Datasets and Evaluation Metrics}

To evaluate the effectiveness of Reasoning-Infused Text Embedding (RITE), we employ the BRIGHT benchmark~\cite{su2024bright}, a collection of 12 retrieval datasets spanning diverse domains such as biology, economics, psychology, robotics, and theoretical disciplines. These datasets are designed to assess retrieval models' ability to handle complex queries requiring logical reasoning and inference.

We measure retrieval performance using Normalized Discounted Cumulative Gain at rank 10 (nDCG@10), a standard metric that captures both relevance ranking quality and retrieval effectiveness.

\vspace{-0.05in}
\subsection{Implementation Details}

We use Mistral 7B and LLaMA 3 8B, two open-source LLMs, as our backbone models. We employ their instruction-tuned variants and set the following hyperparameters for retrieval experiments: temperature to 0, frequency penalty to 0.3, limit the number of response choices to 1, and set maximum input lengths to 256 for passages and 128 for queries. The maximum generation tokens (for reasoning text) are selected from 256, 128, or 64, choosing the value that yields the best retrieval performance for each model. 

For Echo, the embedding is derived from the last-layer hidden states before generating each token in the second occurrence of $x$. While \citet{springer2024repetition} also considered using the last token representation, the mean token representation was found to be more effective in zero-shot scenarios, so we adopt it here. For PR, the embedding is taken from the last-layer hidden state before generating the first output token. Although \citet{zhuang2024promptreps} also experimented with the average over multiple output tokens, we choose the first-token representation for simplicity and efficiency. 

Similarly, for RITE-Echo, we extract the mean token representation of the second occurrence of $x$ as the embedding; for RITE-PR, we extract the hidden state corresponding to the first output token as the embedding, consistent with above.
\vspace{-0.1in}
\section{Results}

\begin{table*}[tb!]
  \small
  \centering
  \caption{Case study of Biology dataset.}
  \vspace{-0.15in}
  \begin{tabular}{p{0.04\linewidth}p{0.84\linewidth}p{0.07\linewidth}}
    \toprule
    \textbf{Type} & \textbf{Text} & \textbf{nDCG@10} \\
    \midrule
    \textbf{Query} & Do animals exhibit handedness (paw-ness?) preference?
    I have been observing my cat and found that when confronted with an unknown item, she will always use her front left paw to touch it.
    This has me wondering if animals exhibit handedness like humans do? (and do I have a left handed cat?)
    One note of importance is that with an unknown item, her approach is always identical, so possibly using the left paw means allowing a fast possible exit based on how she positions her body.
    This question is related to Are there dextral/sinistral higher animals?. However, I question the "paw-ness" as a consequence of how the cat is approaching new items (to be ready to flee), whereas the other question remarks about the high number of "right-pawed" dogs and questions the influence of people for this preference. & \\
    \hline
    \textbf{Oracle} & The question investigates if animals display limb preference, as observed in a cat favoring one paw. The document confirms that limb dominance is common across species, including mammals, birds, and invertebrates, thereby supporting the notion of widespread `paw-ness' in animals. & 0.2372 \\
    \hline
    \textbf{RITE} & To answer your question, a more specific search query would be:
    "Do animals exhibit lateralized behavior or paw preference in performing tasks or interacting with objects?"
    Relevant documents should include:
    * Studies on animal cognition and behavior
    * Research on lateralization in non-human primates, birds, and other animals & 0.8503 \\
    \bottomrule
  \end{tabular}
  \vspace{-0.16in}
  \label{tab:case-study-biology}
\end{table*}

In this section, we analyze the performance of retrieval models across multiple domains using reasoning-infused embeddings. The evaluation considers the impact of reasoning-based infusion on retrieval effectiveness and further compares model-generated reasoning with oracle (human-crafted) reasoning.

\vspace{-0.05in}
\subsection{Impact of Reasoning on Retrieval}

To assess the effectiveness of reasoning-based embeddings, we compare RITE-enhanced methods (RITE-Echo and RITE-PR) against their respective baselines (Echo and PR) using the BRIGHT benchmark, as shown in Table~\ref{tab:results-main}. 
% Table~\ref{tab:results-main} presents the nDCG@10 scores across various domains.

For the Mistral 7B model, we observe a consistent performance boost with RITE-Echo, surpassing Echo across all categories. The most significant improvements include Biology (+58\%), Economics (+18\%), and Programming (Pony) (+71\%). Similarly, RITE-PR shows substantial enhancements over PR, with notable gains in Psychology (+166\%) and Biology (+236\%). These findings indicate that reasoning-based embeddings lead to superior retrieval accuracy by enhancing the contextual understanding of queries, allowing for better alignment with relevant documents.

A similar pattern emerges for the LLaMA 3 8B model, where RITE-Echo improves retrieval effectiveness, particularly in Earth Sciences (+36\%) and Robotics (+258\%). Moreover, RITE-PR significantly enhances retrieval in Psychology (+39\%) and TheoremQA (+353\%). These results reinforce that reasoning-enriched embeddings benefit retrieval in both factual and theoretical domains by bridging the gap between query intent and document relevance.

Additionally, these improvements suggest that reasoning-infused embeddings enhance the model’s ability to handle complex and domain-specific queries. By incorporating reasoning, RITE-based approaches mitigate lexical mismatches and improve semantic coherence in retrieval tasks. This is particularly evident in knowledge-intensive disciplines such as Psychology and Mathematics, where conceptual depth plays a crucial role for effective search.

Overall, reasoning infusion substantially enhances retrieval effectiveness across different knowledge domains, making it a valuable approach for improving search relevance in diverse subject areas. Future advancements could explore fine-tuning reasoning methodologies to further optimize retrieval in specific disciplines.

\vspace{-0.05in}
\subsection{Model-Generated vs. Oracle Reasoning}

We also compare RITE embeddings generated from model-produced reasoning texts (RITE-Echo and RITE-PR) against embeddings using oracle (human-crafted) reasoning texts (GT-Echo and GT-PR). 
% Particularly, we refer to methods using oracle reasoning texts as GT-Echo and GT-PR.
% Specifically, the prompts are: 

% \vspace{0.1in}
% \noindent\fbox{%
%     \centering
%     \parbox{0.98\linewidth}{%
%         \textbf{GT-Echo:} [Oracle reasoning text] Rewrite the query: $x$, rewritten query: $x$
%     }%
% }
% \noindent\fbox{%
%     \centering
%     \parbox{0.98\linewidth}{%
%         \textbf{GT-PR:} Query: $x$. [Oracle reasoning text] Use one most important word to represent the query in a retrieval task. Make sure your word is in lowercase. The word is: ``
%     }%
% }
% \vspace{0.1in}

The results in Tables~\ref{tab:results-oracle-echo} and \ref{tab:results-oracle-pr}, show that oracle reasoning consistently achieves superior retrieval performance. For instance, using the LLaMA 3 8B model, GT-Echo surpasses RITE-Echo in Earth Sciences (+13\%) and TheoremQA (+8\%). Similarly, GT-PR significantly outperforms RITE-PR in TheoremQA (+57\%) and AoPS (+123\%), highlighting the advantages of high-quality, human-authored reasoning texts in capturing domain-specific nuances.

However, RITE-generated reasoning performs competitively in some domains. In Biology and Stack Overflow, the differences between RITE and GT embeddings are minimal, indicating that automatic reasoning generation can approximate human-crafted explanations in specific contexts. This suggests that in fields where explicit domain knowledge is less critical, model-generated reasoning may serve as an efficient alternative to oracle reasoning.

Furthermore, the results indicate that the effectiveness of model-generated reasoning depends on the complexity of the subject matter. In fact-based and structured domains such as programming and biology, RITE performs comparably to human reasoning. Conversely, in abstract and mathematically rigorous areas such as TheoremQA and AoPS, oracle reasoning retains a distinct advantage due to its ability to leverage precise terminologies and formal structures.

These results suggest that while oracle reasoning remains the ideal, model-generated reasoning provides a feasible alternative that substantially enhances retrieval in zero-shot settings. A potential avenue for future work involves improving automatic reasoning techniques by integrating external knowledge bases or leveraging domain-specific language models to refine generated rationales. By doing so, model-generated reasoning could more effectively approximate human-level reasoning across a broader range of disciplines.

\vspace{-0.05in}
\subsection{Case Study Analysis}

We conducted case studies to explore how reasoning-enhanced embeddings influence retrieval effectiveness as an example in Table~\ref{tab:case-study-biology}. 
% Tables~\ref{tab:case-study-biology}, \ref{tab:case-study-theoremqa}, and \ref{tab:case-study-aops} present representative queries and their retrieval outcomes using both oracle and model-generated reasoning.
In the Biology dataset, the RITE-generated reasoning significantly improves retrieval performance for queries related to lateralization in animals, achieving an nDCG@10 of 0.8503 compared to 0.2372 for the oracle reasoning. This improvement stems from the model’s ability to refine queries by emphasizing key biological concepts.

\vspace{-0.1in}
\section{Conclusion}\label{sec:conclusion and Discussion}

In this work, we introduced Reasoning-Infused Text Embedding (RITE), a novel approach that leverages the inherent reasoning capabilities of generative large language models (LLMs) to enhance zero-shot dense retrieval. Our experimental results on the BRIGHT benchmark, which spans diverse domains, demonstrate that RITE consistently surpasses traditional zero-shot embedding methods, such as Echo and PromptReps, by incorporating explicit reasoning into the embedding process.

Our findings highlight that the integration of reasoning significantly enhances retrieval, particularly in tasks requiring complex logical inference. The comparison between model-generated and oracle reasoning texts further underscores the efficacy of RITE, revealing that automatically generated reasoning closely approximates the performance of human-crafted explanations. This indicates that RITE can serve as a powerful tool for improving retrieval effectiveness in applications such as search engines, recommendation systems, and question-answering frameworks.

Additionally, the adaptability of RITE, which supports diverse prompt designs and LLM architectures, underscores its broad applicability across various retrieval scenarios. As generative models continue to advance, further fine-tuning and optimization of RITE have the potential to unlock even greater improvements in retrieval accuracy and computational efficiency. Future work will explore enhancements in prompt engineering, retrieval-specific LLM pretraining, and integration with hybrid retrieval paradigms to further refine and extend the capabilities of RITE.

% In this work, we introduced Reasoning-Infused Text Embedding (RITE), a novel approach that leverages the inherent reasoning capabilities of generative large language models (LLMs) to enhance zero-shot dense retrieval. The experimental results across the BRIGHT benchmark, which encompasses a diverse range of domains, demonstrate that RITE consistently outperforms existing zero-shot language model embedding methods that do not incorporate reasoning, such as Echo and PromptReps.

% Our findings indicate that the inclusion of reasoning significantly improves retrieval performance, particularly in reasoning-intensive tasks. The comparison between model-generated and oracle (ground-truth) reasoning texts further reinforces the effectiveness of RITE, showing that even automatically generated reasoning can approach the performance of ideal human-crafted reasoning. This suggests that RITE can be a valuable tool for enhancing retrieval systems in various applications, including search engines, recommendation systems, and question answering systems. Moreover, the flexibility of the RITE framework, which allows for different prompt designs and LLM backbones, underscores its adaptability to various retrieval contexts. As generative models continue to evolve, we anticipate that further fine-tuning and optimization of RITE could yield even greater improvements in retrieval accuracy and efficiency.

\section{GenAI Usage Disclosure}

During our research, we used a GenAI tool solely for refining the language and improving the clarity of our writing. We did not use it for generating code or for drafting or composing the content of the paper.
% \input{sections/sec7-limitations}

% \begin{acks}
% \red{TODO}
% \end{acks}

\bibliographystyle{ACM-Reference-Format}
\bibliography{anthology,custom}


\begin{thebibliography}{34}

%%% the Web address.

\ifx \showCODEN    \undefined \def \showCODEN     #1{\unskip}     \fi
\ifx \showDOI      \undefined \def \showDOI       #1{#1}\fi
\ifx \showISBNx    \undefined \def \showISBNx     #1{\unskip}     \fi
\ifx \showISBNxiii \undefined \def \showISBNxiii  #1{\unskip}     \fi
\ifx \showISSN     \undefined \def \showISSN      #1{\unskip}     \fi
\ifx \showLCCN     \undefined \def \showLCCN      #1{\unskip}     \fi
\ifx \shownote     \undefined \def \shownote      #1{#1}          \fi
\ifx \showarticletitle \undefined \def \showarticletitle #1{#1}   \fi
\ifx \showURL      \undefined \def \showURL       {\relax}        \fi
% invisible to TeX
\providecommand\bibfield[2]{#2}
\providecommand\bibinfo[2]{#2}
\providecommand\natexlab[1]{#1}
\providecommand\showeprint[2][]{arXiv:#2}

\bibitem[Achiam et~al\mbox{.}(2023)]%
        {achiam2023gpt}
\bibfield{author}{\bibinfo{person}{Josh Achiam}, \bibinfo{person}{Steven Adler}, \bibinfo{person}{Sandhini Agarwal}, \bibinfo{person}{Lama Ahmad}, \bibinfo{person}{Ilge Akkaya}, \bibinfo{person}{Florencia~Leoni Aleman}, \bibinfo{person}{Diogo Almeida}, \bibinfo{person}{Janko Altenschmidt}, \bibinfo{person}{Sam Altman}, \bibinfo{person}{Shyamal Anadkat}, {et~al\mbox{.}}} \bibinfo{year}{2023}\natexlab{}.
\newblock \showarticletitle{Gpt-4 technical report}.
\newblock \bibinfo{journal}{\emph{arXiv preprint arXiv:2303.08774}} (\bibinfo{year}{2023}).
\newblock


\bibitem[BehnamGhader et~al\mbox{.}(2024)]%
        {behnamghader2024llm2vec}
\bibfield{author}{\bibinfo{person}{Parishad BehnamGhader}, \bibinfo{person}{Vaibhav Adlakha}, \bibinfo{person}{Marius Mosbach}, \bibinfo{person}{Dzmitry Bahdanau}, \bibinfo{person}{Nicolas Chapados}, {and} \bibinfo{person}{Siva Reddy}.} \bibinfo{year}{2024}\natexlab{}.
\newblock \showarticletitle{Llm2vec: Large language models are secretly powerful text encoders}.
\newblock \bibinfo{journal}{\emph{arXiv preprint arXiv:2404.05961}} (\bibinfo{year}{2024}).
\newblock


\bibitem[Devlin et~al\mbox{.}(2019)]%
        {devlin-etal-2019-bert}
\bibfield{author}{\bibinfo{person}{Jacob Devlin}, \bibinfo{person}{Ming-Wei Chang}, \bibinfo{person}{Kenton Lee}, {and} \bibinfo{person}{Kristina Toutanova}.} \bibinfo{year}{2019}\natexlab{}.
\newblock \showarticletitle{{BERT}: Pre-training of Deep Bidirectional Transformers for Language Understanding}. In \bibinfo{booktitle}{\emph{Proceedings of the 2019 Conference of the North {A}merican Chapter of the Association for Computational Linguistics: Human Language Technologies, Volume 1 (Long and Short Papers)}}, \bibfield{editor}{\bibinfo{person}{Jill Burstein}, \bibinfo{person}{Christy Doran}, {and} \bibinfo{person}{Thamar Solorio}} (Eds.). \bibinfo{publisher}{Association for Computational Linguistics}, \bibinfo{address}{Minneapolis, Minnesota}, \bibinfo{pages}{4171--4186}.
\newblock
\urldef\tempurl%
\url{https://doi.org/10.18653/v1/N19-1423}
\showDOI{\tempurl}


\bibitem[Isinkaye et~al\mbox{.}(2015)]%
        {isinkaye2015recommendation}
\bibfield{author}{\bibinfo{person}{Folasade~Olubusola Isinkaye}, \bibinfo{person}{Yetunde~O Folajimi}, {and} \bibinfo{person}{Bolande~Adefowoke Ojokoh}.} \bibinfo{year}{2015}\natexlab{}.
\newblock \showarticletitle{Recommendation systems: Principles, methods and evaluation}.
\newblock \bibinfo{journal}{\emph{Egyptian informatics journal}} \bibinfo{volume}{16}, \bibinfo{number}{3} (\bibinfo{year}{2015}), \bibinfo{pages}{261--273}.
\newblock


\bibitem[Ji and Luo(2025)]%
        {ji2025leveraging}
\bibfield{author}{\bibinfo{person}{Cheng Ji} {and} \bibinfo{person}{Huaiying Luo}.} \bibinfo{year}{2025}\natexlab{}.
\newblock \showarticletitle{Leveraging Large Language Model for Intelligent Log Processing and Autonomous Debugging in Cloud AI Platforms}. In \bibinfo{booktitle}{\emph{2025 8th International Conference on Advanced Electronic Materials, Computers and Software Engineering (AEMCSE)}}. IEEE, \bibinfo{pages}{348--351}.
\newblock


\bibitem[Jiang et~al\mbox{.}(2023b)]%
        {jiang2023mistral}
\bibfield{author}{\bibinfo{person}{Albert~Q Jiang}, \bibinfo{person}{Alexandre Sablayrolles}, \bibinfo{person}{Arthur Mensch}, \bibinfo{person}{Chris Bamford}, \bibinfo{person}{Devendra~Singh Chaplot}, \bibinfo{person}{Diego de~las Casas}, \bibinfo{person}{Florian Bressand}, \bibinfo{person}{Gianna Lengyel}, \bibinfo{person}{Guillaume Lample}, \bibinfo{person}{Lucile Saulnier}, {et~al\mbox{.}}} \bibinfo{year}{2023}\natexlab{b}.
\newblock \showarticletitle{Mistral 7B}.
\newblock \bibinfo{journal}{\emph{arXiv preprint arXiv:2310.06825}} (\bibinfo{year}{2023}).
\newblock


\bibitem[Jiang et~al\mbox{.}(2023a)]%
        {jiang2023scaling}
\bibfield{author}{\bibinfo{person}{Ting Jiang}, \bibinfo{person}{Shaohan Huang}, \bibinfo{person}{Zhongzhi Luan}, \bibinfo{person}{Deqing Wang}, {and} \bibinfo{person}{Fuzhen Zhuang}.} \bibinfo{year}{2023}\natexlab{a}.
\newblock \showarticletitle{Scaling sentence embeddings with large language models}.
\newblock \bibinfo{journal}{\emph{arXiv preprint arXiv:2307.16645}} (\bibinfo{year}{2023}).
\newblock


\bibitem[Jin et~al\mbox{.}(2025)]%
        {11010625}
\bibfield{author}{\bibinfo{person}{Yihong Jin}, \bibinfo{person}{Ze Yang}, {and} \bibinfo{person}{Xinhe Xu}.} \bibinfo{year}{2025}\natexlab{}.
\newblock \showarticletitle{Scam Detection for Ethereum Smart Contracts: Leveraging Graph Representation Learning for Secure Blockchain}. In \bibinfo{booktitle}{\emph{2025 4th International Symposium on Computer Applications and Information Technology (ISCAIT)}}. \bibinfo{pages}{1730--1734}.
\newblock
\urldef\tempurl%
\url{https://doi.org/10.1109/ISCAIT64916.2025.11010625}
\showDOI{\tempurl}


\bibitem[Kobayashi and Takeda(2000)]%
        {kobayashi2000information}
\bibfield{author}{\bibinfo{person}{Mei Kobayashi} {and} \bibinfo{person}{Koichi Takeda}.} \bibinfo{year}{2000}\natexlab{}.
\newblock \showarticletitle{Information retrieval on the web}.
\newblock \bibinfo{journal}{\emph{ACM computing surveys (CSUR)}} \bibinfo{volume}{32}, \bibinfo{number}{2} (\bibinfo{year}{2000}), \bibinfo{pages}{144--173}.
\newblock


\bibitem[Lee et~al\mbox{.}(2024)]%
        {lee2024gecko}
\bibfield{author}{\bibinfo{person}{Jinhyuk Lee}, \bibinfo{person}{Zhuyun Dai}, \bibinfo{person}{Xiaoqi Ren}, \bibinfo{person}{Blair Chen}, \bibinfo{person}{Daniel Cer}, \bibinfo{person}{Jeremy~R Cole}, \bibinfo{person}{Kai Hui}, \bibinfo{person}{Michael Boratko}, \bibinfo{person}{Rajvi Kapadia}, \bibinfo{person}{Wen Ding}, {et~al\mbox{.}}} \bibinfo{year}{2024}\natexlab{}.
\newblock \showarticletitle{Gecko: Versatile text embeddings distilled from large language models}.
\newblock \bibinfo{journal}{\emph{arXiv preprint arXiv:2403.20327}} (\bibinfo{year}{2024}).
\newblock


\bibitem[Liu and Chang(2025)]%
        {DBLP:conf/acl/LiuC25}
\bibfield{author}{\bibinfo{person}{Yuxiang Liu} {and} \bibinfo{person}{Kevin~Chen{-}Chuan Chang}.} \bibinfo{year}{2025}\natexlab{}.
\newblock \showarticletitle{Writing Like the Best: Exemplar-Based Expository Text Generation}. In \bibinfo{booktitle}{\emph{Proceedings of the 63rd Annual Meeting of the Association for Computational Linguistics (Volume 1: Long Papers), {ACL} 2025, Vienna, Austria, July 27 - August 1, 2025}}. \bibinfo{publisher}{Association for Computational Linguistics}, \bibinfo{pages}{25739--25764}.
\newblock
\urldef\tempurl%
\url{https://aclanthology.org/2025.acl-long.1250/}
\showURL{%
\tempurl}


\bibitem[Liu et~al\mbox{.}(2023)]%
        {DBLP:conf/emnlp/Liu0C23}
\bibfield{author}{\bibinfo{person}{Yuxiang Liu}, \bibinfo{person}{Jie Huang}, {and} \bibinfo{person}{Kevin~Chen{-}Chuan Chang}.} \bibinfo{year}{2023}\natexlab{}.
\newblock \showarticletitle{Ask To The Point: Open-Domain Entity-Centric Question Generation}. In \bibinfo{booktitle}{\emph{Findings of the Association for Computational Linguistics: {EMNLP} 2023, Singapore, December 6-10, 2023}}. \bibinfo{publisher}{Association for Computational Linguistics}, \bibinfo{pages}{2703--2716}.
\newblock
\urldef\tempurl%
\url{https://doi.org/10.18653/v1/2023.findings-emnlp.178}
\showURL{%
\tempurl}


\bibitem[Luo and Ji(2025a)]%
        {luo2025cross}
\bibfield{author}{\bibinfo{person}{Huaiying Luo} {and} \bibinfo{person}{Cheng Ji}.} \bibinfo{year}{2025}\natexlab{a}.
\newblock \showarticletitle{Cross-Cloud Data Privacy Protection: Optimizing Collaborative Mechanisms of AI Systems by Integrating Federated Learning and LLMs}.
\newblock \bibinfo{journal}{\emph{arXiv preprint arXiv:2505.13292}} (\bibinfo{year}{2025}).
\newblock


\bibitem[Luo and Ji(2025b)]%
        {luo2025federated}
\bibfield{author}{\bibinfo{person}{Huaiying Luo} {and} \bibinfo{person}{Cheng Ji}.} \bibinfo{year}{2025}\natexlab{b}.
\newblock \showarticletitle{Federated Learning-Based Data Collaboration Method for Enhancing Edge Cloud AI System Security Using Large Language Models}.
\newblock \bibinfo{journal}{\emph{arXiv preprint arXiv:2506.18087}} (\bibinfo{year}{2025}).
\newblock


\bibitem[Ma et~al\mbox{.}(2023a)]%
        {ma-etal-2023-query}
\bibfield{author}{\bibinfo{person}{Xinbei Ma}, \bibinfo{person}{Yeyun Gong}, \bibinfo{person}{Pengcheng He}, \bibinfo{person}{Hai Zhao}, {and} \bibinfo{person}{Nan Duan}.} \bibinfo{year}{2023}\natexlab{a}.
\newblock \showarticletitle{Query Rewriting in Retrieval-Augmented Large Language Models}. In \bibinfo{booktitle}{\emph{Proceedings of the 2023 Conference on Empirical Methods in Natural Language Processing}}, \bibfield{editor}{\bibinfo{person}{Houda Bouamor}, \bibinfo{person}{Juan Pino}, {and} \bibinfo{person}{Kalika Bali}} (Eds.). \bibinfo{publisher}{Association for Computational Linguistics}, \bibinfo{address}{Singapore}, \bibinfo{pages}{5303--5315}.
\newblock
\urldef\tempurl%
\url{https://doi.org/10.18653/v1/2023.emnlp-main.322}
\showDOI{\tempurl}


\bibitem[Ma et~al\mbox{.}(2023b)]%
        {ma2023fine}
\bibfield{author}{\bibinfo{person}{Xueguang Ma}, \bibinfo{person}{Liang Wang}, \bibinfo{person}{Nan Yang}, \bibinfo{person}{Furu Wei}, {and} \bibinfo{person}{Jimmy Lin}.} \bibinfo{year}{2023}\natexlab{b}.
\newblock \showarticletitle{Fine-tuning llama for multi-stage text retrieval}.
\newblock \bibinfo{journal}{\emph{arXiv preprint arXiv:2310.08319}} (\bibinfo{year}{2023}).
\newblock


\bibitem[Muennighoff et~al\mbox{.}(2024)]%
        {muennighoff2024generative}
\bibfield{author}{\bibinfo{person}{Niklas Muennighoff}, \bibinfo{person}{Hongjin Su}, \bibinfo{person}{Liang Wang}, \bibinfo{person}{Nan Yang}, \bibinfo{person}{Furu Wei}, \bibinfo{person}{Tao Yu}, \bibinfo{person}{Amanpreet Singh}, {and} \bibinfo{person}{Douwe Kiela}.} \bibinfo{year}{2024}\natexlab{}.
\newblock \showarticletitle{Generative representational instruction tuning}.
\newblock \bibinfo{journal}{\emph{arXiv preprint arXiv:2402.09906}} (\bibinfo{year}{2024}).
\newblock


\bibitem[Patil et~al\mbox{.}(2023)]%
        {patil2023survey}
\bibfield{author}{\bibinfo{person}{Rajvardhan Patil}, \bibinfo{person}{Sorio Boit}, \bibinfo{person}{Venkat Gudivada}, {and} \bibinfo{person}{Jagadeesh Nandigam}.} \bibinfo{year}{2023}\natexlab{}.
\newblock \showarticletitle{A survey of text representation and embedding techniques in nlp}.
\newblock \bibinfo{journal}{\emph{IEEE Access}}  \bibinfo{volume}{11} (\bibinfo{year}{2023}), \bibinfo{pages}{36120--36146}.
\newblock


\bibitem[Springer et~al\mbox{.}(2024)]%
        {springer2024repetition}
\bibfield{author}{\bibinfo{person}{Jacob~Mitchell Springer}, \bibinfo{person}{Suhas Kotha}, \bibinfo{person}{Daniel Fried}, \bibinfo{person}{Graham Neubig}, {and} \bibinfo{person}{Aditi Raghunathan}.} \bibinfo{year}{2024}\natexlab{}.
\newblock \showarticletitle{Repetition improves language model embeddings}.
\newblock \bibinfo{journal}{\emph{arXiv preprint arXiv:2402.15449}} (\bibinfo{year}{2024}).
\newblock


\bibitem[Su et~al\mbox{.}(2024)]%
        {su2024bright}
\bibfield{author}{\bibinfo{person}{Hongjin Su}, \bibinfo{person}{Howard Yen}, \bibinfo{person}{Mengzhou Xia}, \bibinfo{person}{Weijia Shi}, \bibinfo{person}{Niklas Muennighoff}, \bibinfo{person}{Han-yu Wang}, \bibinfo{person}{Haisu Liu}, \bibinfo{person}{Quan Shi}, \bibinfo{person}{Zachary~S Siegel}, \bibinfo{person}{Michael Tang}, {et~al\mbox{.}}} \bibinfo{year}{2024}\natexlab{}.
\newblock \showarticletitle{BRIGHT: A Realistic and Challenging Benchmark for Reasoning-Intensive Retrieval}.
\newblock \bibinfo{journal}{\emph{arXiv preprint arXiv:2407.12883}} (\bibinfo{year}{2024}).
\newblock


\bibitem[Trivedi et~al\mbox{.}(2023)]%
        {trivedi-etal-2023-interleaving}
\bibfield{author}{\bibinfo{person}{Harsh Trivedi}, \bibinfo{person}{Niranjan Balasubramanian}, \bibinfo{person}{Tushar Khot}, {and} \bibinfo{person}{Ashish Sabharwal}.} \bibinfo{year}{2023}\natexlab{}.
\newblock \showarticletitle{Interleaving Retrieval with Chain-of-Thought Reasoning for Knowledge-Intensive Multi-Step Questions}. In \bibinfo{booktitle}{\emph{Proceedings of the 61st Annual Meeting of the Association for Computational Linguistics (Volume 1: Long Papers)}}, \bibfield{editor}{\bibinfo{person}{Anna Rogers}, \bibinfo{person}{Jordan Boyd-Graber}, {and} \bibinfo{person}{Naoaki Okazaki}} (Eds.). \bibinfo{publisher}{Association for Computational Linguistics}, \bibinfo{address}{Toronto, Canada}, \bibinfo{pages}{10014--10037}.
\newblock
\urldef\tempurl%
\url{https://doi.org/10.18653/v1/2023.acl-long.557}
\showDOI{\tempurl}


\bibitem[Wang et~al\mbox{.}(2024b)]%
        {wang2024blendfilter}
\bibfield{author}{\bibinfo{person}{Haoyu Wang}, \bibinfo{person}{Tuo Zhao}, {and} \bibinfo{person}{Jing Gao}.} \bibinfo{year}{2024}\natexlab{b}.
\newblock \showarticletitle{BlendFilter: Advancing Retrieval-Augmented Large Language Models via Query Generation Blending and Knowledge Filtering}.
\newblock \bibinfo{journal}{\emph{arXiv preprint arXiv:2402.11129}} (\bibinfo{year}{2024}).
\newblock


\bibitem[Wang et~al\mbox{.}(2022)]%
        {wang2022text}
\bibfield{author}{\bibinfo{person}{Liang Wang}, \bibinfo{person}{Nan Yang}, \bibinfo{person}{Xiaolong Huang}, \bibinfo{person}{Binxing Jiao}, \bibinfo{person}{Linjun Yang}, \bibinfo{person}{Daxin Jiang}, \bibinfo{person}{Rangan Majumder}, {and} \bibinfo{person}{Furu Wei}.} \bibinfo{year}{2022}\natexlab{}.
\newblock \showarticletitle{Text embeddings by weakly-supervised contrastive pre-training}.
\newblock \bibinfo{journal}{\emph{arXiv preprint arXiv:2212.03533}} (\bibinfo{year}{2022}).
\newblock


\bibitem[Wang et~al\mbox{.}(2024a)]%
        {wang-etal-2024-improving-text}
\bibfield{author}{\bibinfo{person}{Liang Wang}, \bibinfo{person}{Nan Yang}, \bibinfo{person}{Xiaolong Huang}, \bibinfo{person}{Linjun Yang}, \bibinfo{person}{Rangan Majumder}, {and} \bibinfo{person}{Furu Wei}.} \bibinfo{year}{2024}\natexlab{a}.
\newblock \showarticletitle{Improving Text Embeddings with Large Language Models}. In \bibinfo{booktitle}{\emph{Proceedings of the 62nd Annual Meeting of the Association for Computational Linguistics (Volume 1: Long Papers)}}, \bibfield{editor}{\bibinfo{person}{Lun-Wei Ku}, \bibinfo{person}{Andre Martins}, {and} \bibinfo{person}{Vivek Srikumar}} (Eds.). \bibinfo{publisher}{Association for Computational Linguistics}, \bibinfo{address}{Bangkok, Thailand}, \bibinfo{pages}{11897--11916}.
\newblock
\urldef\tempurl%
\url{https://doi.org/10.18653/v1/2024.acl-long.642}
\showDOI{\tempurl}


\bibitem[Xiao et~al\mbox{.}(2024)]%
        {xiao2024rar}
\bibfield{author}{\bibinfo{person}{Chenghao Xiao}, \bibinfo{person}{G~Thomas Hudson}, {and} \bibinfo{person}{Noura~Al Moubayed}.} \bibinfo{year}{2024}\natexlab{}.
\newblock \showarticletitle{RAR-b: Reasoning as Retrieval Benchmark}.
\newblock \bibinfo{journal}{\emph{arXiv preprint arXiv:2404.06347}} (\bibinfo{year}{2024}).
\newblock


\bibitem[Xu et~al\mbox{.}(2024a)]%
        {10.1145/3627673.3679071}
\bibfield{author}{\bibinfo{person}{Han Xu}, \bibinfo{person}{Xingyuan Wang}, {and} \bibinfo{person}{Haipeng Chen}.} \bibinfo{year}{2024}\natexlab{a}.
\newblock \showarticletitle{Towards Real-Time and Personalized Code Generation}. In \bibinfo{booktitle}{\emph{Proceedings of the 33rd ACM International Conference on Information and Knowledge Management}}. \bibinfo{pages}{5568–5569}.
\newblock


\bibitem[Xu et~al\mbox{.}(2024b)]%
        {xu2024can}
\bibfield{author}{\bibinfo{person}{Han Xu}, \bibinfo{person}{Jingyang Ye}, \bibinfo{person}{Yutong Li}, {and} \bibinfo{person}{Haipeng Chen}.} \bibinfo{year}{2024}\natexlab{b}.
\newblock \showarticletitle{Can Speculative Sampling Accelerate ReAct Without Compromising Reasoning Quality?}. In \bibinfo{booktitle}{\emph{The Second Tiny Papers Track at ICLR 2024}}.
\newblock
\urldef\tempurl%
\url{https://openreview.net/forum?id=42b9hJrIpX}
\showURL{%
\tempurl}


\bibitem[Xu et~al\mbox{.}(2024c)]%
        {xu2024restful}
\bibfield{author}{\bibinfo{person}{Han Xu}, \bibinfo{person}{Ruining Zhao}, \bibinfo{person}{Jindong Wang}, {and} \bibinfo{person}{Haipeng Chen}.} \bibinfo{year}{2024}\natexlab{c}.
\newblock \showarticletitle{RESTful-Llama: Connecting User Queries to RESTful APIs}. In \bibinfo{booktitle}{\emph{Proceedings of the 2024 Conference on Empirical Methods in Natural Language Processing: Industry Track}}. \bibinfo{pages}{1433--1443}.
\newblock


\bibitem[Yang et~al\mbox{.}(2025)]%
        {11065413}
\bibfield{author}{\bibinfo{person}{Ze Yang}, \bibinfo{person}{Yihong Jin}, \bibinfo{person}{Juntian Liu}, \bibinfo{person}{Xinhe Xu}, \bibinfo{person}{Yihan Zhang}, {and} \bibinfo{person}{Shuyang Ji}.} \bibinfo{year}{2025}\natexlab{}.
\newblock \showarticletitle{Research on Cloud Platform Network Traffic Monitoring and Anomaly Detection System based on Large Language Models}. In \bibinfo{booktitle}{\emph{2025 IEEE 7th International Conference on Communications, Information System and Computer Engineering (CISCE)}}. \bibinfo{pages}{1029--1032}.
\newblock
\urldef\tempurl%
\url{https://doi.org/10.1109/CISCE65916.2025.11065413}
\showDOI{\tempurl}


\bibitem[Yang et~al\mbox{.}(2024)]%
        {yang2024hades}
\bibfield{author}{\bibinfo{person}{Ze Yang}, \bibinfo{person}{Yihong Jin}, {and} \bibinfo{person}{Xinhe Xu}.} \bibinfo{year}{2024}\natexlab{}.
\newblock \showarticletitle{Hades: Hardware accelerated decoding for efficient speculation in large language models}.
\newblock \bibinfo{journal}{\emph{arXiv preprint arXiv:2412.19925}} (\bibinfo{year}{2024}).
\newblock


\bibitem[Yu et~al\mbox{.}(2023)]%
        {yu2023chain}
\bibfield{author}{\bibinfo{person}{Wenhao Yu}, \bibinfo{person}{Hongming Zhang}, \bibinfo{person}{Xiaoman Pan}, \bibinfo{person}{Kaixin Ma}, \bibinfo{person}{Hongwei Wang}, {and} \bibinfo{person}{Dong Yu}.} \bibinfo{year}{2023}\natexlab{}.
\newblock \showarticletitle{Chain-of-note: Enhancing robustness in retrieval-augmented language models}.
\newblock \bibinfo{journal}{\emph{arXiv preprint arXiv:2311.09210}} (\bibinfo{year}{2023}).
\newblock


\bibitem[Zhang et~al\mbox{.}(2024a)]%
        {zhang2024simple}
\bibfield{author}{\bibinfo{person}{Bowen Zhang}, \bibinfo{person}{Kehua Chang}, {and} \bibinfo{person}{Chunping Li}.} \bibinfo{year}{2024}\natexlab{a}.
\newblock \showarticletitle{Simple techniques for enhancing sentence embeddings in generative language models}. In \bibinfo{booktitle}{\emph{International Conference on Intelligent Computing}}. Springer, \bibinfo{pages}{52--64}.
\newblock


\bibitem[Zhang et~al\mbox{.}(2024b)]%
        {zhang-etal-2024-end}
\bibfield{author}{\bibinfo{person}{Jiahao Zhang}, \bibinfo{person}{Haiyang Zhang}, \bibinfo{person}{Dongmei Zhang}, \bibinfo{person}{Liu Yong}, {and} \bibinfo{person}{Shen Huang}.} \bibinfo{year}{2024}\natexlab{b}.
\newblock \showarticletitle{End-to-End Beam Retrieval for Multi-Hop Question Answering}. In \bibinfo{booktitle}{\emph{Proceedings of the 2024 Conference of the North American Chapter of the Association for Computational Linguistics: Human Language Technologies (Volume 1: Long Papers)}}, \bibfield{editor}{\bibinfo{person}{Kevin Duh}, \bibinfo{person}{Helena Gomez}, {and} \bibinfo{person}{Steven Bethard}} (Eds.). \bibinfo{publisher}{Association for Computational Linguistics}, \bibinfo{address}{Mexico City, Mexico}, \bibinfo{pages}{1718--1731}.
\newblock
\urldef\tempurl%
\url{https://doi.org/10.18653/v1/2024.naacl-long.96}
\showDOI{\tempurl}


\bibitem[Zhuang et~al\mbox{.}(2024)]%
        {zhuang2024promptreps}
\bibfield{author}{\bibinfo{person}{Shengyao Zhuang}, \bibinfo{person}{Xueguang Ma}, \bibinfo{person}{Bevan Koopman}, \bibinfo{person}{Jimmy Lin}, {and} \bibinfo{person}{Guido Zuccon}.} \bibinfo{year}{2024}\natexlab{}.
\newblock \showarticletitle{PromptReps: Prompting Large Language Models to Generate Dense and Sparse Representations for Zero-Shot Document Retrieval}.
\newblock \bibinfo{journal}{\emph{arXiv preprint arXiv:2404.18424}} (\bibinfo{year}{2024}).
\newblock


\end{thebibliography}

% \appendix

\end{document}